\documentclass[acmlarge,nonacm]{asrlart}
\usepackage{graphicx} % Required for inserting images
\usepackage[nolist]{acronym} %nohyperlinks
\usepackage{xcolor}
\usepackage[hyperfootnotes=false]{hyperref}
\hypersetup{colorlinks, linkcolor={red!50!black}, citecolor={blue!50!black}, urlcolor={blue!50!black}}
\usepackage{todonotes}
\usepackage{booktabs} %for hrules in tables
\usepackage{tabularx} % for newcolumntype
\usepackage{placeins} %floatbarrier
\usepackage[toc, page]{appendix}
\usepackage{soul}
\usepackage{pgffor}
\usepackage{siunitx}
\usepackage{amsmath}

\AtBeginDocument{%
  \providecommand\BibTeX{{%
    \normalfont B\kern-0.5em{\scshape i\kern-0.25em b}\kern-0.8em\TeX}}}

\acrodef{UWB}[UWB]{ultra-wideband}
\acrodef{mocap}[mocap]{motion capture}
\acrodef{ROS}[ROS]{Robot Operating System}
\acrodef{CAD}[CAD]{Computer Aided Design}
\acrodef{LOS}[LOS]{line-of-sight}
\acrodef{NLOS}[NLOS]{none-line-of-sight}
\acrodef{GPU}[GPU]{Graphical Processing Unit}
\acrodef{SLAM}[SLAM]{Simultaneous Localization and Mapping}
\acrodef{IMU}[IMU]{inertial measurement unit}

\newcommand{\frames}[1]{\ensuremath{\mathcal{F}_{#1}}}

\renewcommand{\i}[1]{\textit{#1}}

\newcolumntype{C}[1]{>{\centering\let\newline\\\arraybackslash\hspace{0pt}}m{#1}}
\newcolumntype{L}[1]{>{\raggedright\let\newline\\\arraybackslash\hspace{0pt}}m{#1}}
\newcolumntype{R}[1]{>{\raggedleft\let\newline\\\arraybackslash\hspace{0pt}}m{#1}}

\def\apriltag{\textit{Apriltag}}
\def\apriltags{\textit{Apriltags}}

\begin{document}
\settopmatter{printacmref=false}

\author{Frederike D\"umbgen}
\authornote{The authors have contributed equally to this work. In particular, the UWB processing was provided by CC and MS, and the stereo-camera processing was provided by FD, CH. The authors collected the data together at University of Toronto.}
\affiliation{%
  \institution{University of Toronto}
  \city{Toronto}
}
\author{Mohammed A. Shalaby}
\authornotemark[1]
\affiliation{%
  \institution{McGill University}
  \city{Montreal}
}
\author{Connor Holmes}
\authornotemark[1]
\affiliation{%
  \institution{University of Toronto}
  \city{Toronto}
}
\author{Charles C. Cossette}
\authornotemark[1]
\affiliation{%
  \institution{McGill University}
  \city{Montreal}
}
\author{James R. Forbes}
\affiliation{%
  \institution{McGill University}
  \city{Montreal}
}
\author{Jerome Le Ny}
\affiliation{%
  \institution{Polytechnique Montreal}
  \city{Montreal}
}
\author{Timothy D. Barfoot}
\affiliation{%
  \institution{University of Toronto}
  \city{Toronto}
}

\renewcommand{\shortauthors}{STAR-loc}

\title{STAR-loc: Dataset for STereo And Range-based localization}

%\includeonly{_sections/data-overview}

%\include{_sections/abstract}
\begin{abstract}
    This document contains a detailed description of the STAR-loc dataset. For a quick starting guide please refer to the associated Github repository (https://github.com/utiasASRL/starloc). The dataset consists of stereo camera data (rectified/raw images and inertial measurement unit measurements) and ultra-wideband (UWB) data (range measurements) collected on a sensor rig in a Vicon motion capture arena. The UWB anchors and visual landmarks (Apriltags) are of known position, so the dataset can be used for both localization and Simultaneous Localization and Mapping (SLAM).

\end{abstract}

\maketitle

%\blfootnote{$^*$ The authors have contributed equally to the work. In particular, the~\ac{UWB} hardware and processing was provided by CC, MS, JN and JF, and the stereo-camera hardware and processing was provided by FD, CH, TB.}

\setlength{\parskip}{.5\baselineskip}

\vspace{-0.5em}

\section{Summary}\label{sec:summary}

This dataset contains multiple trajectories of a custom sensor rig inside a \ac{mocap} arena. Attached to the sensor rig are a stereo camera, providing image streams and \ac{IMU} data, and one or two \ac{UWB} tags, providing distance measurements to up to 8 fixed and known \ac{UWB} anchors. Also fixed in the \ac{mocap} arena are 55 \i{Apriltag}~\cite{apriltag} landmarks of known position, which are detected online by the stereo camera. A depiction of the experimental setup can be found in Figure~\ref{fig:starloc}. 

A compact version of the dataset can be found on \href{https://github.com/utiasASRL/starloc/}{Github} and is the recommended starting point for using the dataset. It contains all pre-processed \i{csv} files of the data, as well as some convenience functions for reading the data. If required, the full dataset can be found on \href{https://drive.google.com/drive/folders/1vcDn_rwebtP1KaJjTYqWjZCvtu1W_EnB?usp=share_link}{Google Drive}. In addition to the aforementioned \i{csv} files, the full dataset also includes the raw \i{bag files} and many analysis plots, making it easier to choose which part of the dataset to use for the given application. Both the Github repository and drive are structured as follows:

\begin{figure}[ht]
\centering
\includegraphics[width=.7\linewidth]{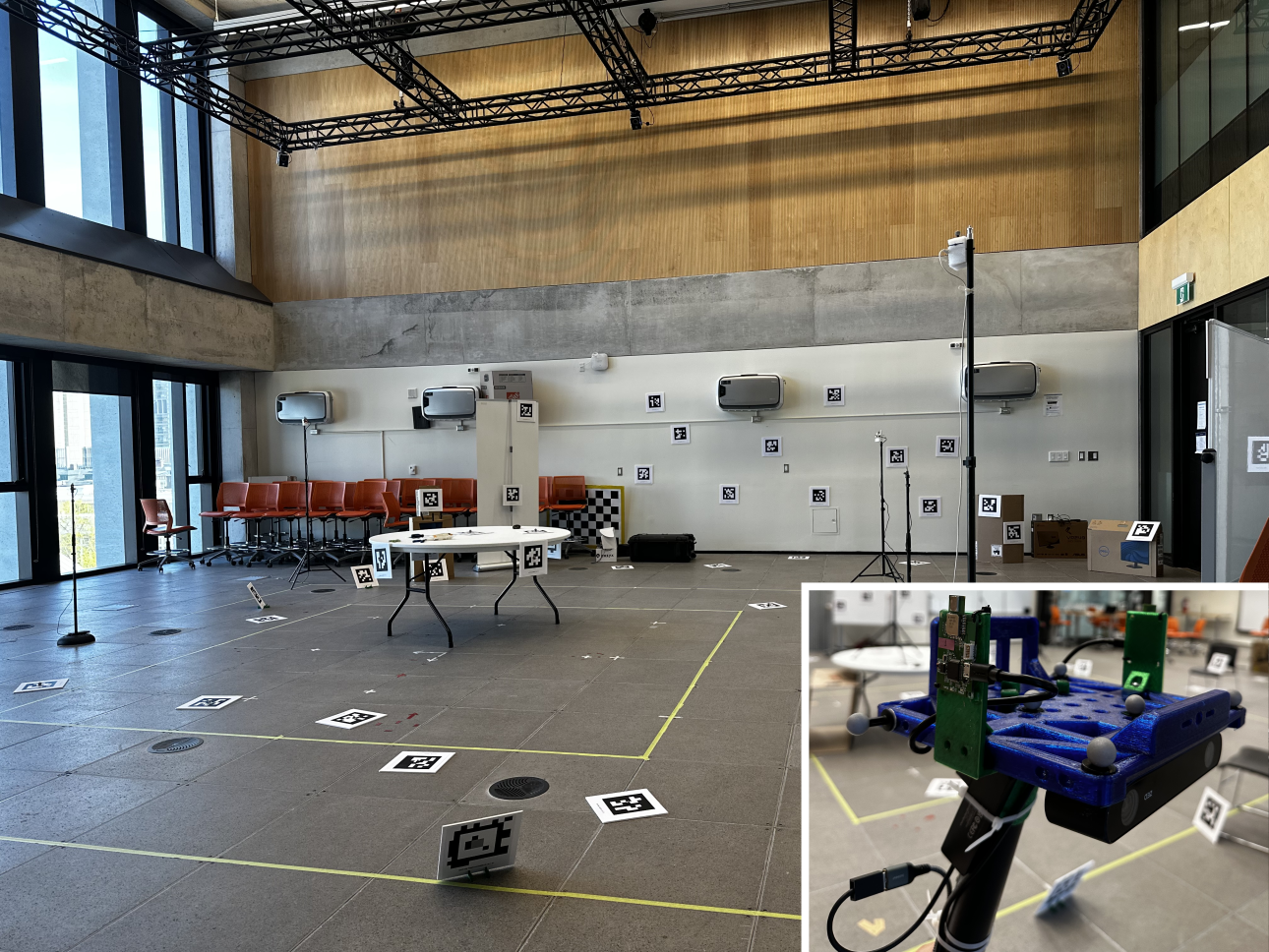}
\caption{The STAR-loc dataset contains data collected on different trajectories of a sensor rig (bottom right) in a Vicon \ac{mocap} arena, providing range and calibrated stereo measurements.}
\label{fig:starloc}
\end{figure}

\begin{itemize}
    \item \texttt{data/<dataset-name>/}: folder containing all data of the run of name \texttt{<dataset-name>} (see Section~\ref{sec:data-overview} for an overview of the different runs). 
    \begin{itemize}
        \item csv files of processed data (\texttt{apriltag.csv}, \texttt{imu.csv}, \texttt{uwb.csv}, \texttt{gt.csv}. See \texttt{data/README.md} for information on the data fields.
        \item \texttt{calib.json}: file with calibration information of this run.
        \item (Drive only) bag files of this run (\texttt{<dataset-name>\_0.db3} and \texttt{metadata.yaml})
        \item (Drive only) some analysis plots of the data, see Appendix~\ref{app:data} for  information on the plots.
        \item (Drive only) the video of the left camera with overlaid Apriltag detections.
    \end{itemize}
  \item \texttt{mocap/}: folder containg environment information. 
    \begin{itemize}
      \item \texttt{uwb\_markers\_<version>.csv},\texttt{mocap/environment\_<version>.csv}: csv files of the \ac{UWB} anchor and \i{Apriltag} landmark locations, respectively, used in Setup of \texttt{<version>}. See Section~\ref{sec:data-overview} for descriptions of the different setup versions.
      \item (Drive only) plots of the different setup versions.
      \item (Drive only) vsk and tag config files used in calibration procedure.
    \end{itemize}
\end{itemize}

%\clearpage\FloatBarrier
\section{System description}\label{sec:system}

\begin{figure}[h]
  \centering
  \begin{minipage}{.4\textwidth}
  \centering
  \includegraphics[width=\linewidth, angle=270]{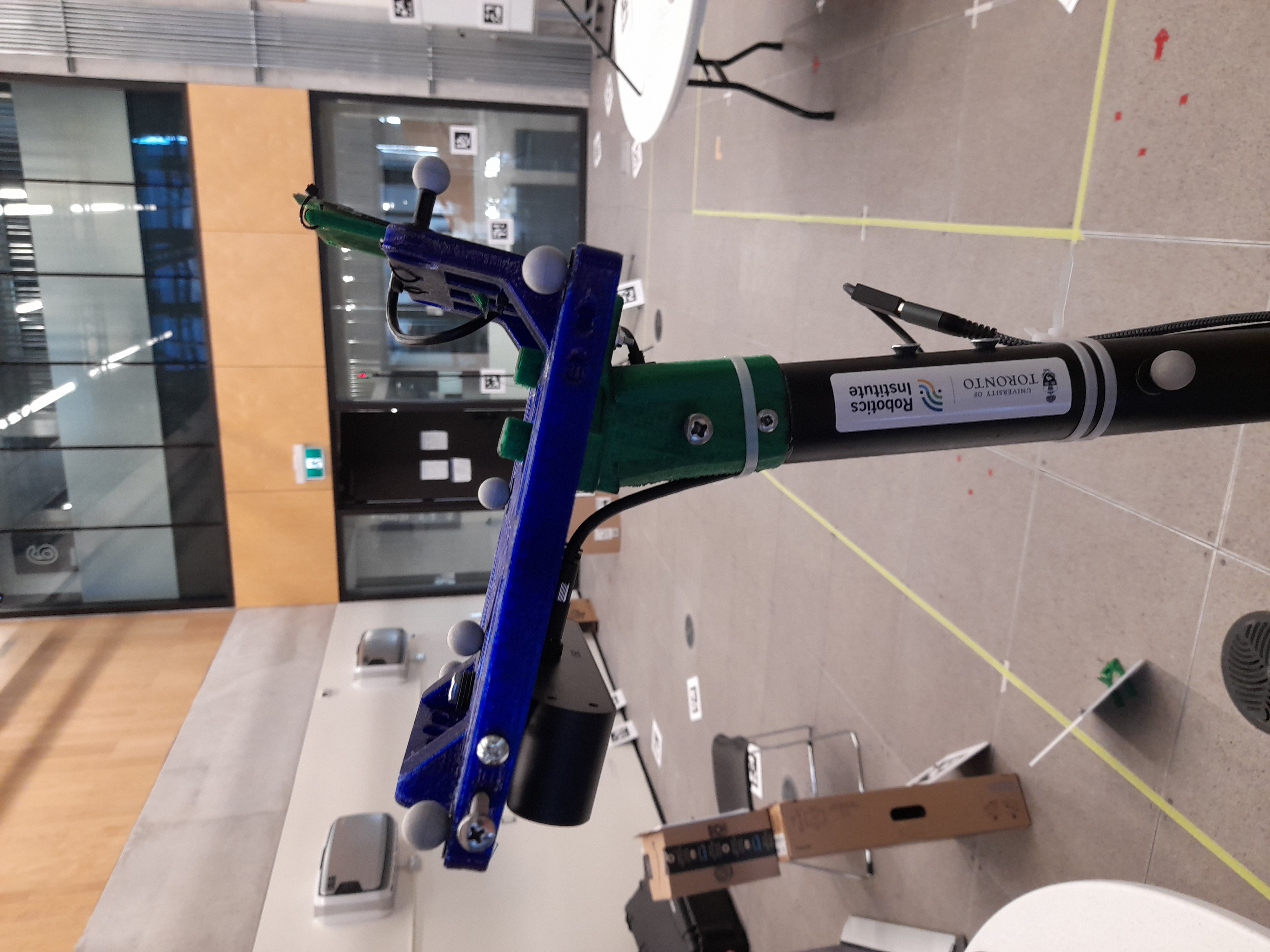}
  \end{minipage}
  \begin{minipage}{.49\textwidth}
  \centering
  \includegraphics[width=\linewidth]{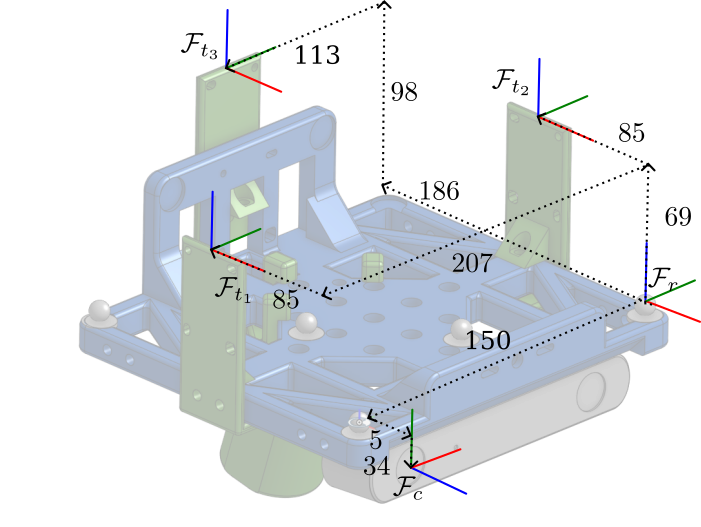}
  \end{minipage}
  \caption{Left: the used sensor rig with one \ac{UWB} tag and stereo camera attached. Right: the corresponding \ac{CAD} model including the definitions of the sensor frames, showing the three possible \ac{UWB} tag locations \frames{t_1}, \frames{t_2} and \frames{t_3} and the stereo camera frame \frames{c}.}
  \label{fig:sensorrig}
\end{figure}

A photo of the sensor rig used for data collection is shown in Figure~\ref{fig:sensorrig}. Attached to the rig are the \ac{UWB} tag(s) and the stereo camera. A laptop \footnote{\i{Lenovo P16 Thinkpad} with 16GB RAM and \i{RTXA4500} \i{NVIDIA} \i{GPU}} (not shown) is used for driving all sensors and collecting the data through the help of dedicated \ac{ROS} nodes. 

Figure~\ref{fig:sensorrig} also depicts the different sensor frames including their transforms. The frames are:
\begin{itemize}
  \item \frames{r}: robot frame (tracked by \ac{mocap} system)
  \item \frames{c}: camera frame
  \item \frames{t_i}: $i$-th tag frame, with $i=1\ldots 3$
\end{itemize}

\subsection{\ac{UWB} tag and anchors}

The main~\ac{UWB} system used are custom-made boards provided by the Mobile Robotics and Autonomous Systems Laboratory (MRASL), where each board is fitted with DWM1000~\ac{UWB} transceivers\footnote{Available at \url{https://www.qorvo.com/products/p/DWM1000}.}. The board is shown in Figure~\ref{fig:landmarks}. These boards are equipped with a STM32F405RG microcontroller\footnote{Available at \url{https://www.st.com/en/microcontrollers-microprocessors/stm32f405rg.html}.} to interface with the DWM1000 modules, and the boards are then connected to the onboard computer using USB. Details about the ranging and communication protocols can be found in~\cite{uwb-hardware}. We also provide a small dataset with the off-the-shelf development kit MDEK1001\footnote{Available at \url{https://www.qorvo.com/products/p/MDEK1001}.} (setup \texttt{s5}). We use custom data collection scripts to record distance measurements. 

\begin{figure}[htb]
  \begin{minipage}{.49\textwidth}
  \centering
  \includegraphics[width=5cm,angle=270]{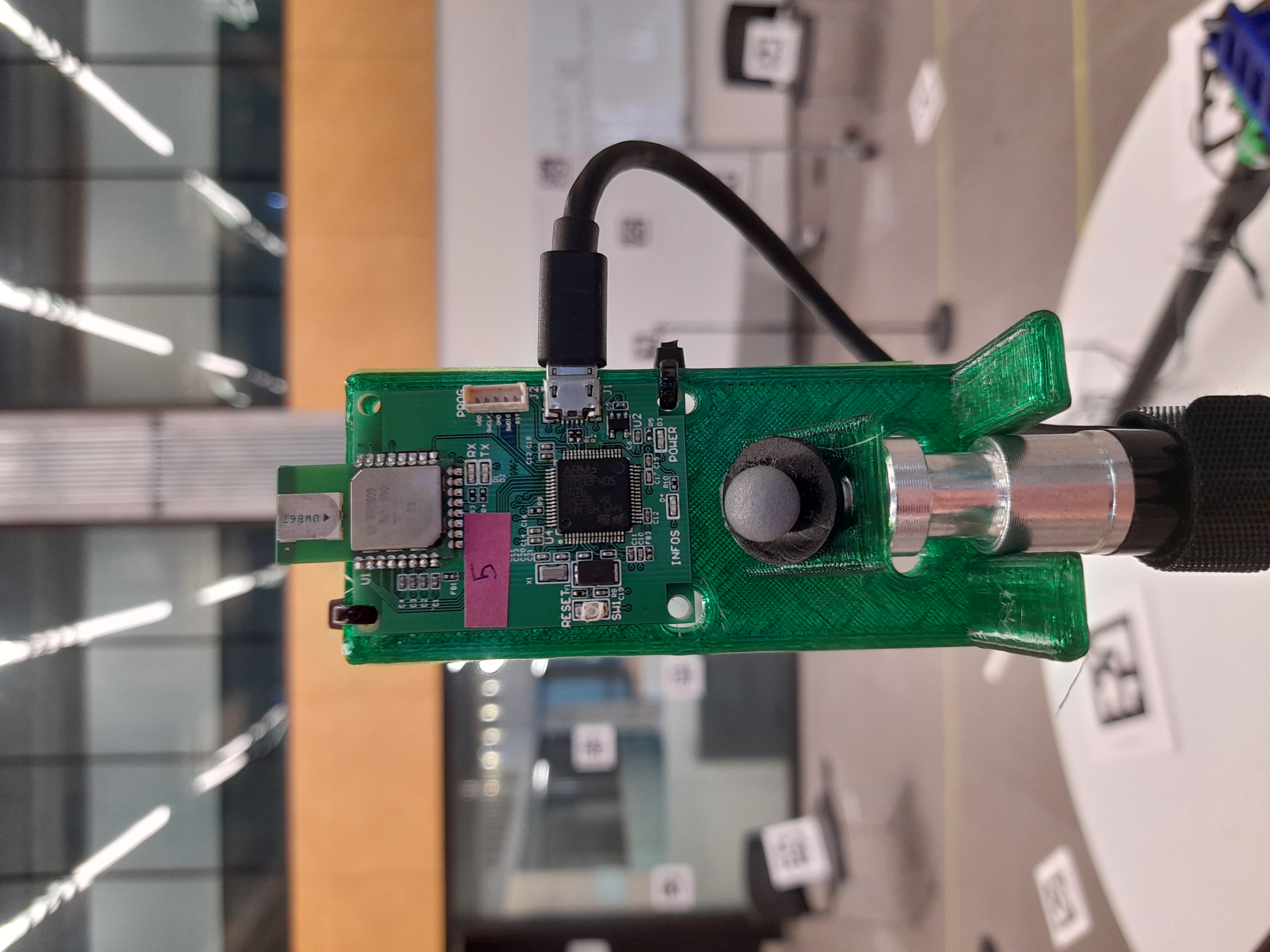}
  \end{minipage}
  \begin{minipage}{.49\textwidth}
  \includegraphics[height=5cm]{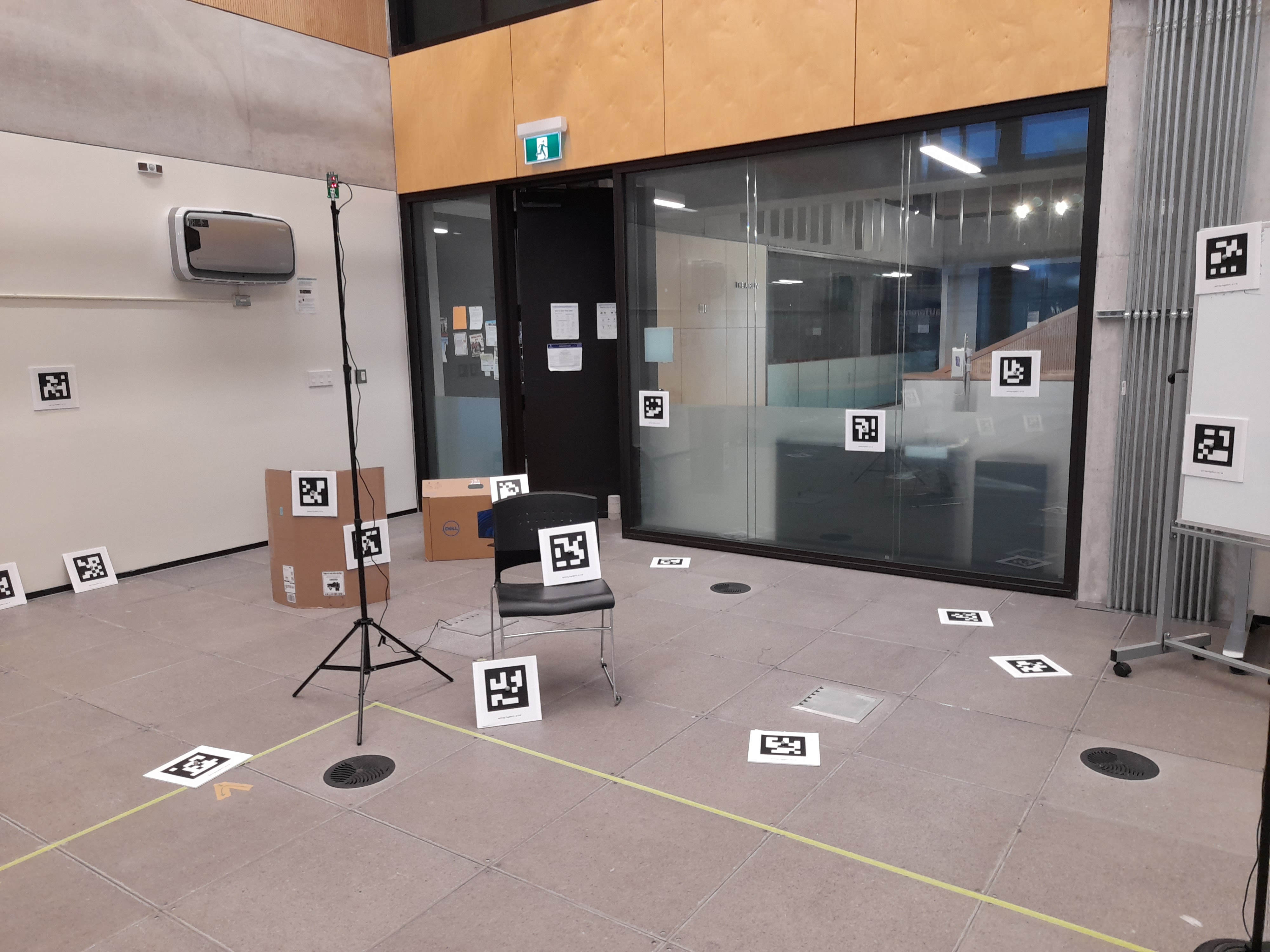}
  \end{minipage}
  \caption{Left: One of eight fixed \ac{UWB} anchors. Right: subset of \apriltag~landmarks.}
  \label{fig:landmarks}
\end{figure}

\subsection{Visual landmarks}

We use \apriltags~\cite{apriltag}, tag version \i{Tag36h11}, as known landmarks.  The tags are printed on US letter paper, glued to foam board, and spread throughout the room at fixed locations of different heights and orientations, as shown in Figure~\ref{fig:landmarks}.  We use the open-source libraries~\href{https://github.com/christianrauch/apriltag_ros}{apriltag\_ros} and~\href{https://github.com/christianrauch/apriltag_msg}{apriltag\_msg} for \apriltag~detection.

\subsection{Stereo camera}

We use the \i{ZED2i} stereo camera from \i{stereolabs} with a \SI{12}{cm} baseline and set to $1920\times1080$ resolution.
For data collection, we employ the \ac{ROS}-based camera driver provided by stereolabs, using the \href{https://github.com/ros-perception/image_transport_plugins}{image\_transport\_plugins} \ac{ROS} packages for image compression.

\subsection{Motion capture}

We use a \i{VICON} \ac{mocap} system for ground truth monitoring. For data collection, we use the~\href{https://github.com/OPT4SMART/ros2-vicon-receiver}{ros2-vicon-receiver} package by \i{OPT4SMART} to publish the native data stream from \i{VICON} as \ac{ROS} messages.

%\clearpage\FloatBarrier
\section{Dataset overview}\label{sec:data-overview}

\begin{table}[ht]
  \centering
  \caption{Overview of datasets}\label{tab:overview}
  \begin{tabular}{lcccccL{7cm}}
\toprule
\textbf{setup}     & \i{s1} & \i{s2} & \i{s3} & \i{s4} & \i{s5} \\
\textbf{landmarks} & \i{v1} & \i{v1} & \i{v2} & \i{v2} & \i{v3} \\
\textbf{trajectory} & & & & &  & description \\
\midrule
loop-2d       & x & x & x & x & x & multiple loops, keeping sensor rig level\\
loop-2d-fast  & x & x & x &   &   & multiple loops, walking fast, keeping sensor rig level\\
loop-3d       & x & x & x &   & x & multiple loops, pointing sensor rig at different directions \\
zigzag        &   & x & x & x &   & a roughly piecewise linear trajectory, keeping sensor rig level \\
eight         &   & x & x &   &   & draw large eights in the air, pointing at an area dense in \apriltags\\
\midrule
apriltag      &   &   & x &   &   & one loop with complete \i{Apriltag} detections \\
ell           &   &   & x &   &   & multiple ell-shaped trajectories with a lot of height variation \\
grid          &   &   & x &   &   & static measurements at 8 grid points, including one full circle at last point. \\
loop-3d-z     &   &   & x &   &   & loop with a lot of height variation \\
\bottomrule
\end{tabular}
\end{table}

\subsection{Runs and setups}
An overview of all collected datasets is shown in Table~\ref{tab:overview}. The datasets are distinguished by their setup, landmarks and trajectory version. These aspects are described in more details in the paragraphs below. 

\paragraph{Trajectory versions}

\begin{figure}[ht]
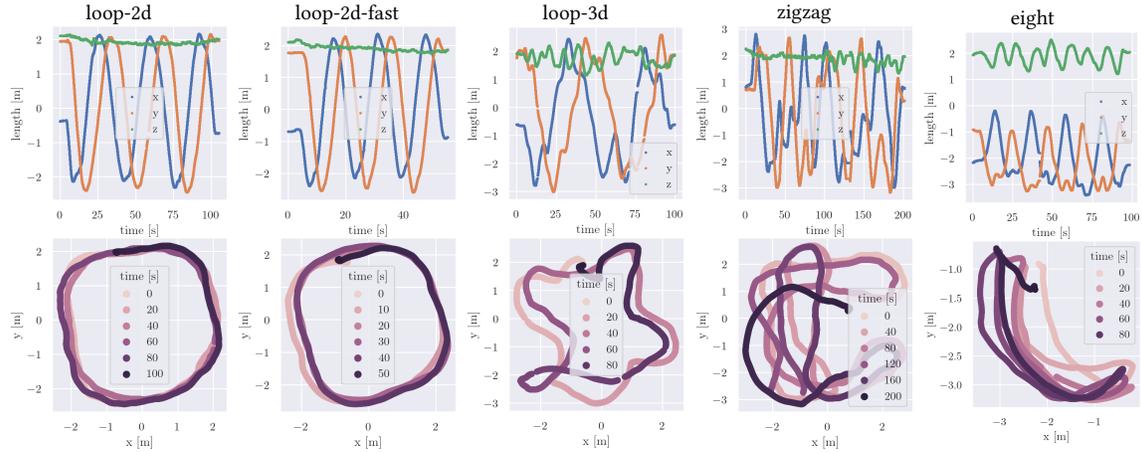

  \centering
  \foreach \name in {%
  loop-2d,
  loop-2d-fast,
  loop-3d,
  zigzag,
  eight%
  }{%
  \edef\filename{\name_s3}
  \begin{minipage}{.18\linewidth}\centering \footnotesize{\name}\\
  \includegraphics[width=\linewidth]{_figures/\filename/gt-t-xyz.png} 
  \includegraphics[width=\linewidth]{_figures/\filename/gt-xy.png} 
  \end{minipage}
  }
  \caption{Overview of the different standard trajectory types recorded with many different setups. The shown data corresponds to setup \textit{s3}.}\label{fig:overview-standard}
\end{figure}

\begin{figure}[ht]
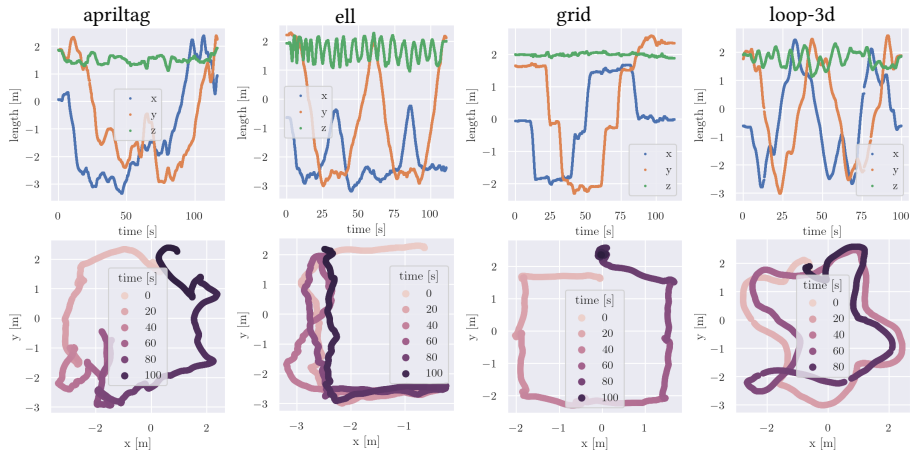

  \centering
  \foreach \name in {%
  apriltag,
  ell,
  grid,
  loop-3d%
  }{%
  \edef\filename{\name_s3}
  \begin{minipage}{.18\linewidth}\centering \footnotesize{\name}\\
  \includegraphics[width=\linewidth]{_figures/\filename/gt-t-xyz.png} 
  \includegraphics[width=\linewidth]{_figures/\filename/gt-xy.png} 
  \end{minipage}
  }
  \caption{Overview of the trajectory types recorded with the setup \textit{s3}.}\label{fig:overview-more}
\end{figure}

The first five listed trajectory types (also denoted ``standard'' in what follows) are repeated for different kinds of tag setups to facilitate their comparison. The four remaining datasets are recorded for setup \textit{s3} only, and serve as datasets to highlight specific phenomena.  Example trajectories for the 5 standard dataset types are shown in Figure~\ref{fig:overview-standard}.

An overview of the remaining datasets is shown in Figure~\ref{fig:overview-more}.

\paragraph{Setup versions}
The datasets include 5 different setups of the sensor rig:
\begin{itemize}
  \item \i{s1}: We use locations 1 and 2 (see Figure~\ref{fig:sensorrig}) for the \ac{UWB} tags 1 and 2, respectively, on the sensor rig. The rig is held just above the person's height using a single \i{PVC} tube.
  \item \i{s2}: hand-held up, 2 tags: The sensor rig is mounted on a second \i{PVC} tube to increase the proportion of \ac{LOS} regions: the rig is now a half metre above the person's height.
  \item \i{s3}: hand-held up, 1 tag: We use only location 3 (see Figure~\ref{fig:sensorrig}) for the \ac{UWB} tag 1. 
  \item \i{s4}: mounted on Jackal robot, 1 tag at location 3.
  \item \i{s5}: hand-held up, 1 Decawave tag at location 3. 
\end{itemize}

\paragraph{Landmarks versions}
The datasets include three different landmark layouts. The first two differ only minimally in geometry; Version \i{v2} is an improved setup of \i{v1} after analyzing noise patterns of both the \ac{mocap} system and the \ac{UWB} anchors. Version \i{v3} is the layout used for \i{Decawave} data and consists of only four anchors. The different layouts are shown Figure~\ref{fig:landmarks}. 
\begin{figure}[h]
  \centering
  \begin{minipage}{.3\linewidth} \centering
    v1 \\
  \includegraphics[width=\linewidth]{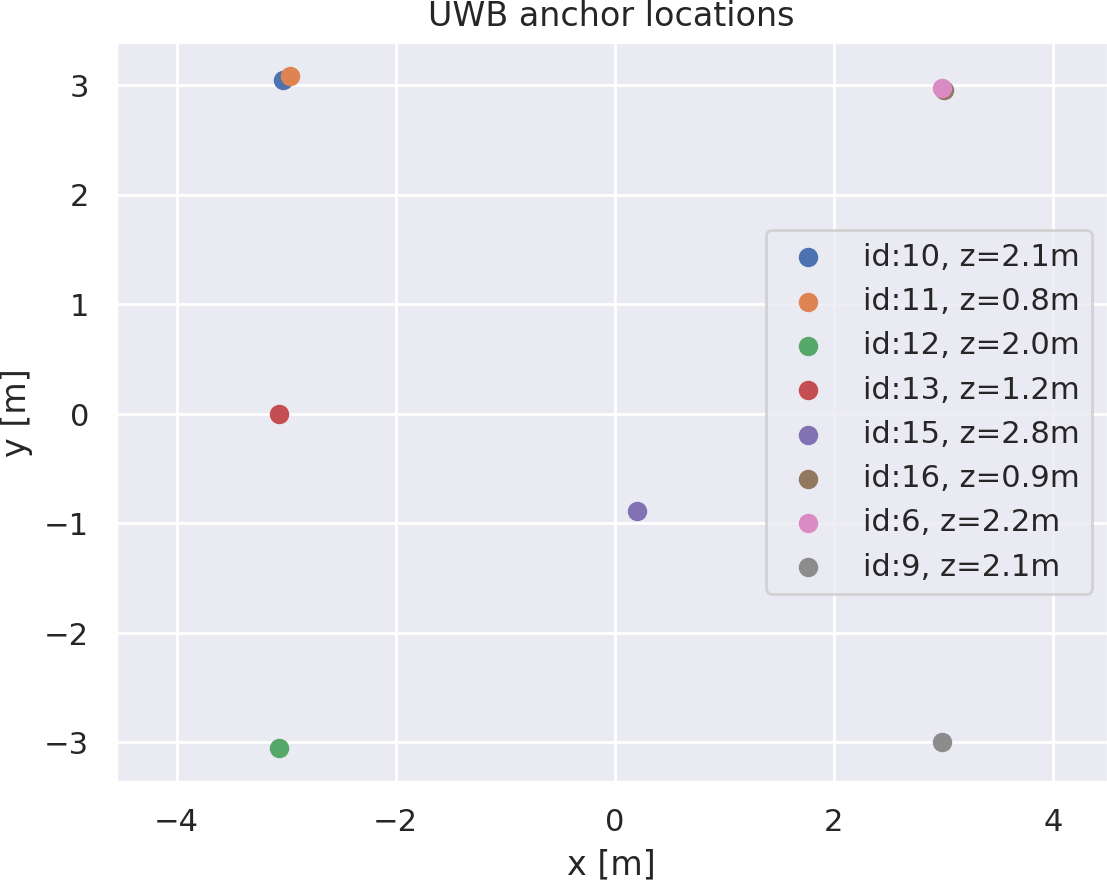}
  \includegraphics[width=\linewidth]{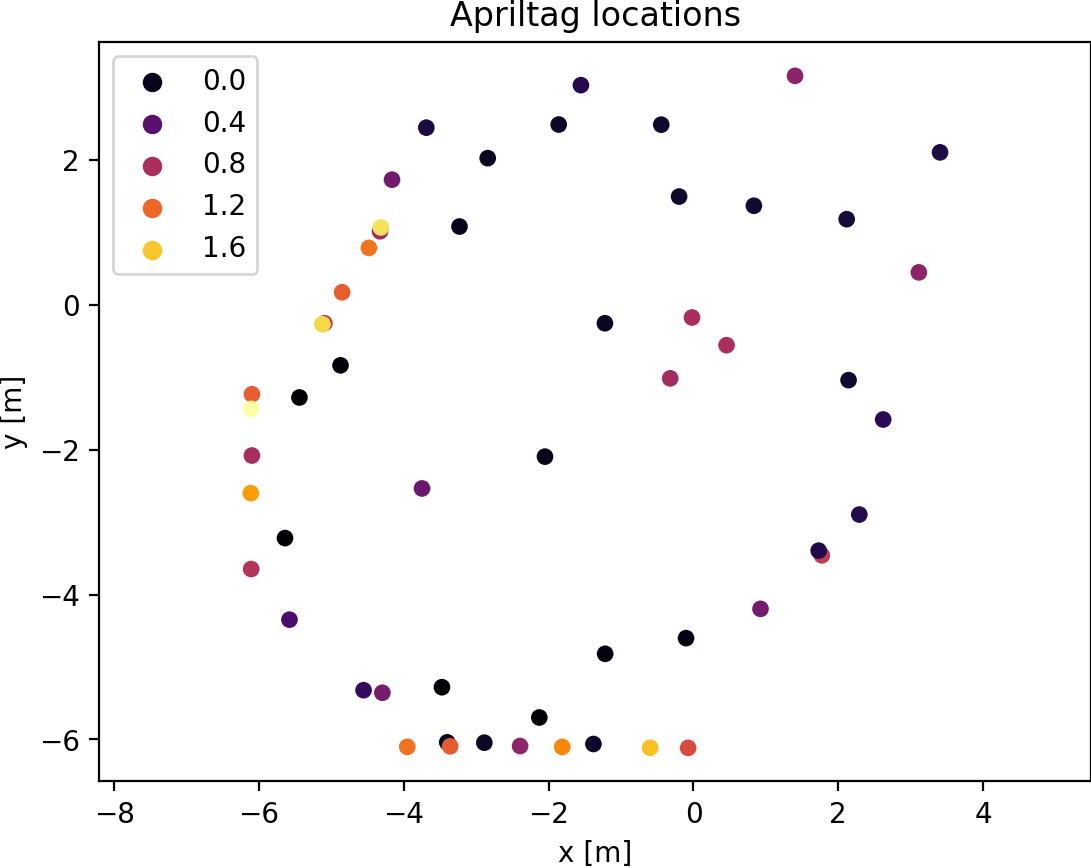}
  \end{minipage}
  \begin{minipage}{.3\linewidth} \centering
    v2 \\
  \includegraphics[width=\linewidth]{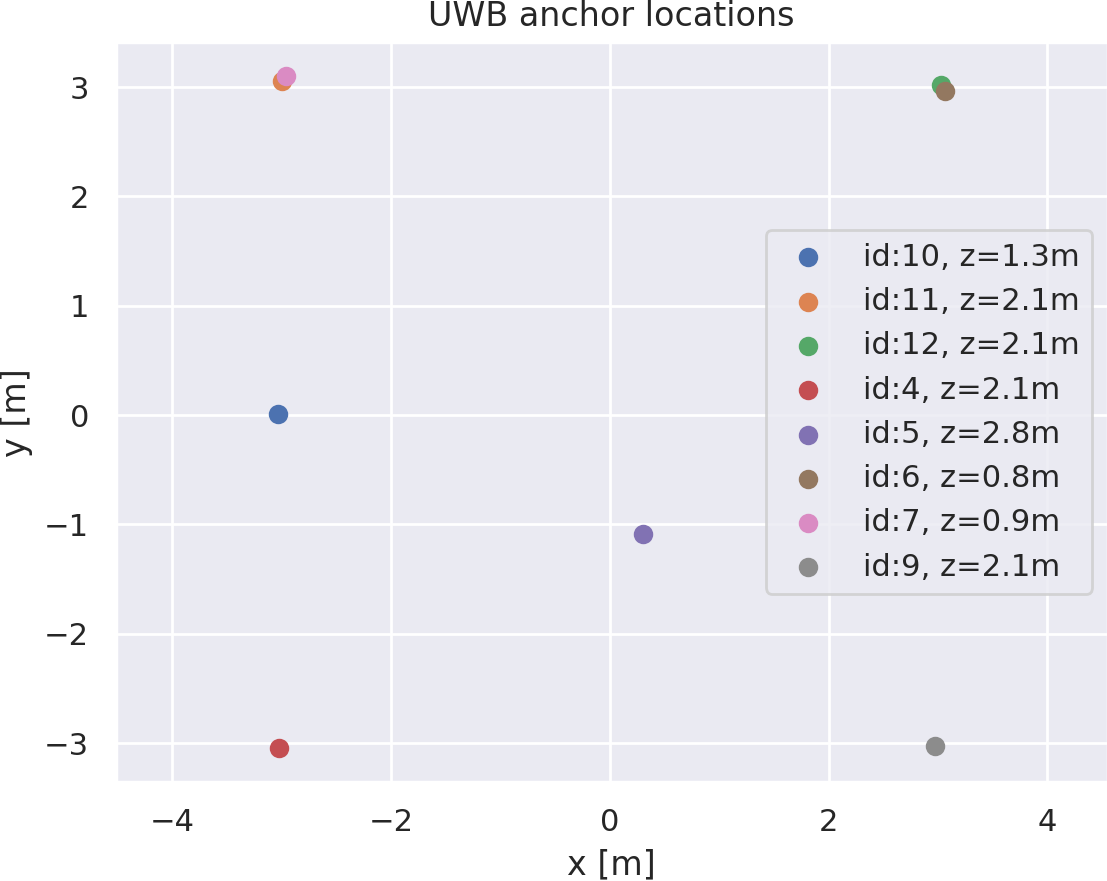}
  \includegraphics[width=\linewidth]{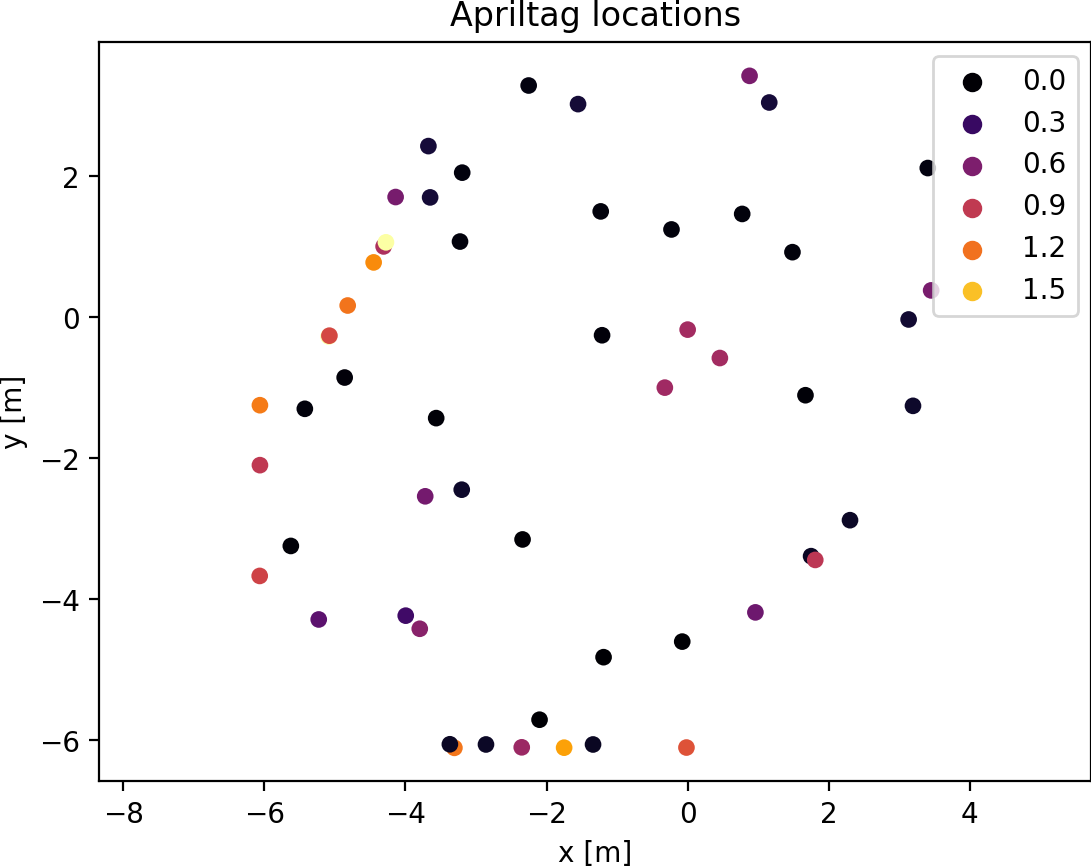}
  \end{minipage}
  \begin{minipage}{.3\linewidth} \centering
    v3 \\
  \includegraphics[width=\linewidth]{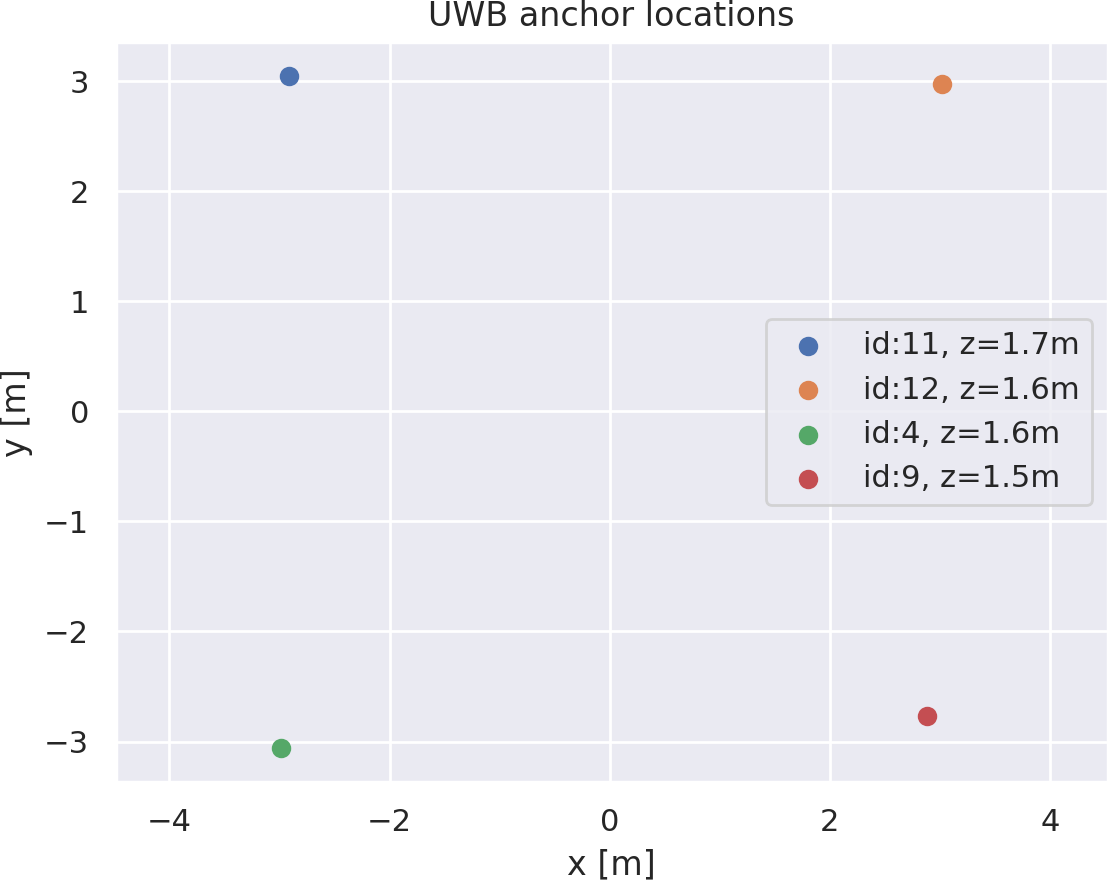}
  \end{minipage}
  \caption{The three different \ac{UWB} anchor layouts and two different \i{Apriltag} landmark layouts used.}
  \label{fig:landmarks}
\end{figure}

\newpage
\subsection{Collected data}

In this Section, we report some statistics taken over all the collected datasets. An individual analysis per dataset can be found in Appendix~\ref{app:data}. The names of the datasets are of the format:
\begin{equation*}
  \texttt{trajectory-version\_setup-version}.
\end{equation*}

\paragraph{Number of messages and rates}
The collected data rates and number of messages received, for each dataset, are shown in Table~\ref{tab:overview-rates}. 

\begin{table}[h]
\centering
\caption{Overview of all collected datasets} \label{tab:overview-rates}
\small{
\begin{tabular}{lccccr}
\toprule
& \multicolumn{4}{c}{median rate [Hz]} & duration [s] \\
data type & Apriltag & UWB & Vicon & IMU &  \\
\midrule
loop-2d\_s1 & 14.9 & 266.1 & 48.0 & 193.8 & 65.6 \\
loop-2d-fast\_s1 & 14.9 & 260.7 & 48.0 & 194.1 & 34.2 \\
loop-3d\_s1 & 14.9 & 249.3 & 48.0 & 193.6 & 87.7 \\
loop-2d\_s2 & 14.9 & 251.3 & 48.0 & 193.3 & 86.9 \\
loop-2d-fast\_s2 & 14.9 & 246.4 & 48.1 & 192.8 & 35.3 \\
loop-3d\_s2 & 14.9 & 244.6 & 48.0 & 192.6 & 80.3 \\
zigzag\_s2 & 14.9 & 256.4 & 48.0 & 193.0 & 78.0 \\
eight\_s2 & 15.0 & 246.6 & 48.1 & 192.3 & 49.2 \\
loop-2d\_s3 & 14.9 & 248.6 & 48.0 & 192.9 & 105.3 \\
loop-2d-fast\_s3 & 14.9 & 247.8 & 48.0 & 193.1 & 55.5 \\
loop-3d\_s3 & 14.9 & 249.4 & 48.0 & 193.9 & 99.8 \\
zigzag\_s3 & 14.9 & 249.4 & 48.0 & 193.9 & 201.6 \\
eight\_s3 & 14.9 & 247.4 & 48.0 & 192.9 & 98.5 \\
loop-2d\_s4 & 11.3 & 247.4 & 48.3 & 196.2 & 194.1 \\
zigzag\_s4 & 14.3 & 249.1 & 48.0 & 198.0 & 218.2 \\
grid\_s3 & 14.9 & 248.6 & 48.0 & 192.7 & 112.9 \\
ell\_s3 & 14.9 & 248.7 & 48.0 & 192.6 & 111.1 \\
loop-3d-z\_s3 & 14.9 & 249.1 & 48.0 & 193.2 & 150.3 \\
apriltag\_s3 & 14.9 & 247.8 & 48.0 & 193.1 & 116.4 \\
loop-2d\_s5 & 14.9 & 40.0 & 48.0 & 193.2 & 82.0 \\
loop-2d-v2\_s5 & 14.9 & 40.0 & 48.0 & 193.6 & 101.0 \\
loop-3d\_s5 & 14.9 & 40.0 & 48.0 & 193.1 & 67.0 \\
\midrule
median rates [Hz] & 14.9 & 248.6 & 48.0 & 193.2 &  \\
\bottomrule
\end{tabular}

}
\end{table}

\begin{figure}[h]
  \centering
  \includegraphics[height=4cm]{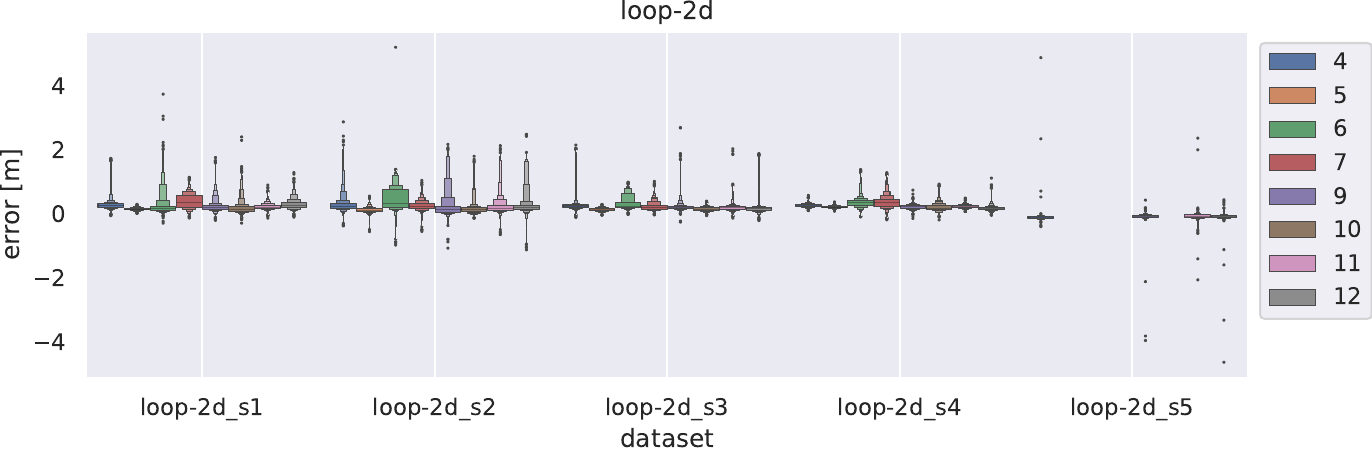}
  \includegraphics[height=4cm]{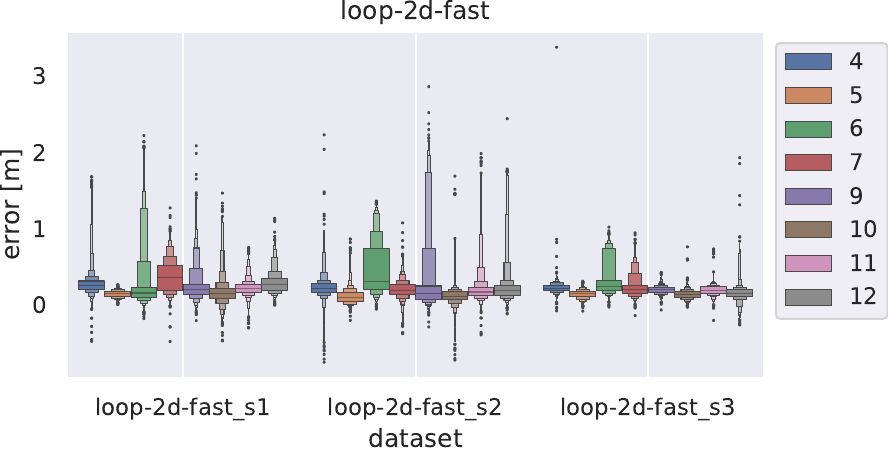}
  \includegraphics[height=4cm]{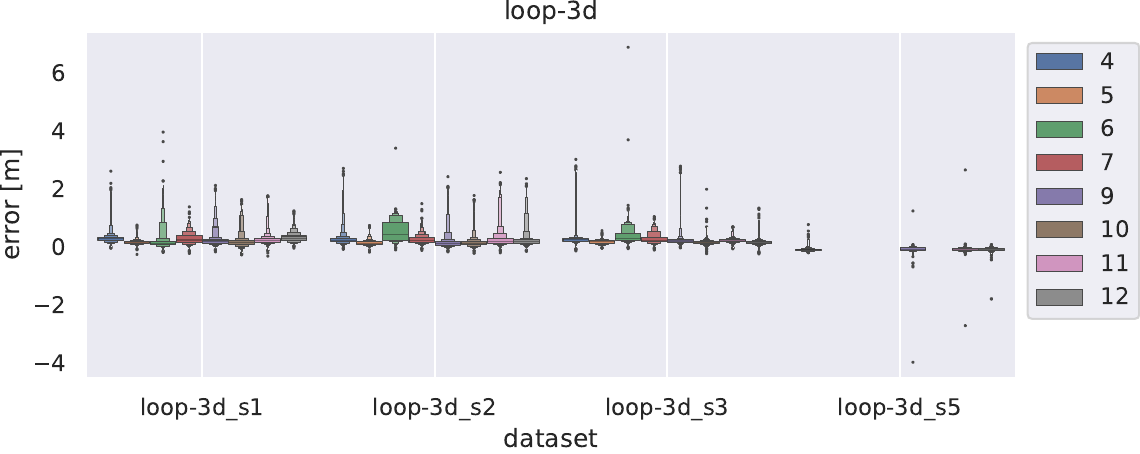}
  \includegraphics[height=4cm]{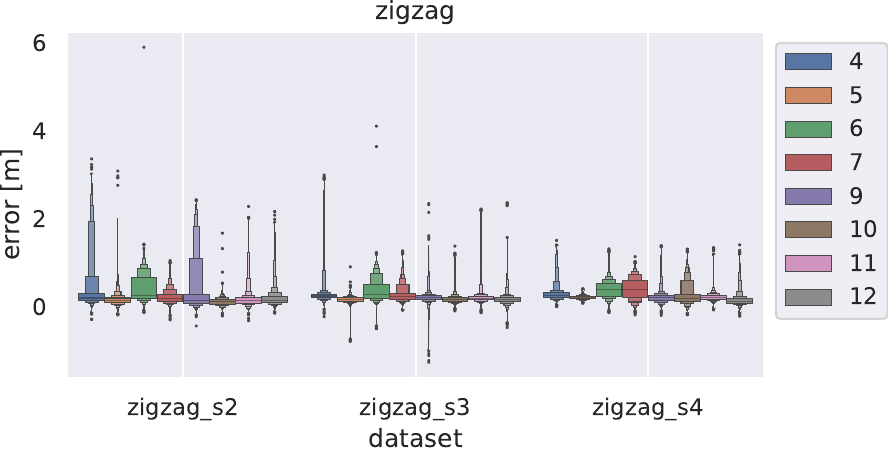}
  \includegraphics[height=4cm]{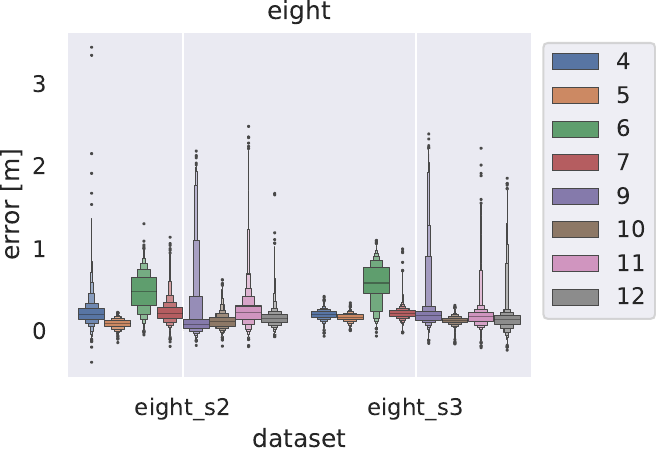}
  \includegraphics[height=4cm]{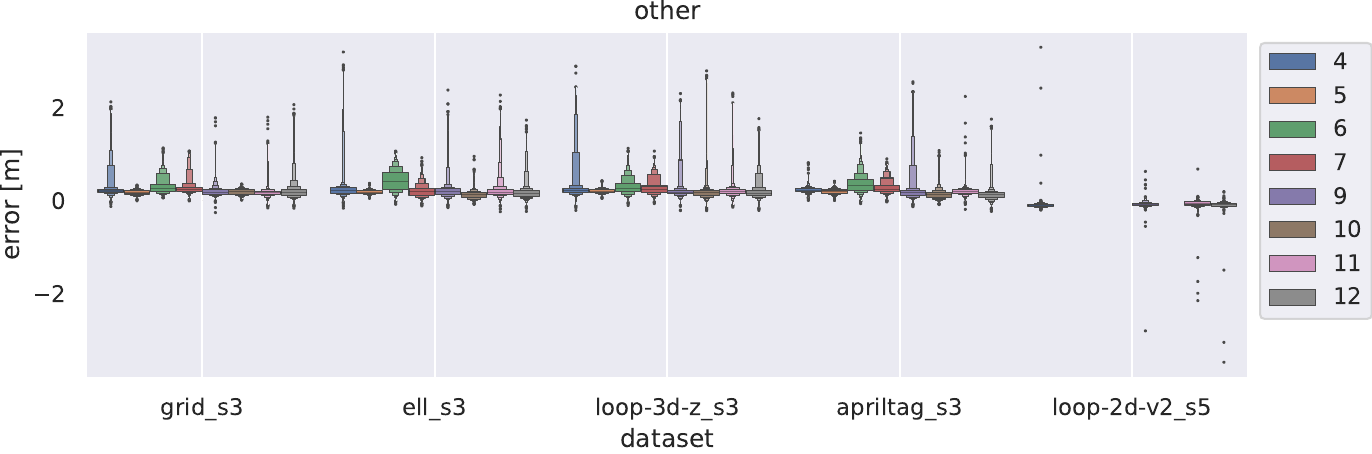}
  \caption{\ac{UWB} performance over all datasets.}
  \label{fig:overview-uwb}
\end{figure}

\paragraph{\ac{UWB} quality}
The quality of \ac{UWB} measurements per dataset, measured by the (signed) difference between the ground truth and measured distances, is shown in Figure~\ref{fig:overview-uwb}. The measurements shown are \textit{before calibration}. For performance after calibration, please refer to the individual dataset plots provided in the dataset on Google Drive.

\paragraph{Stereo quality}

\begin{figure}[h]
  \centering
  \includegraphics[height=3.9cm]{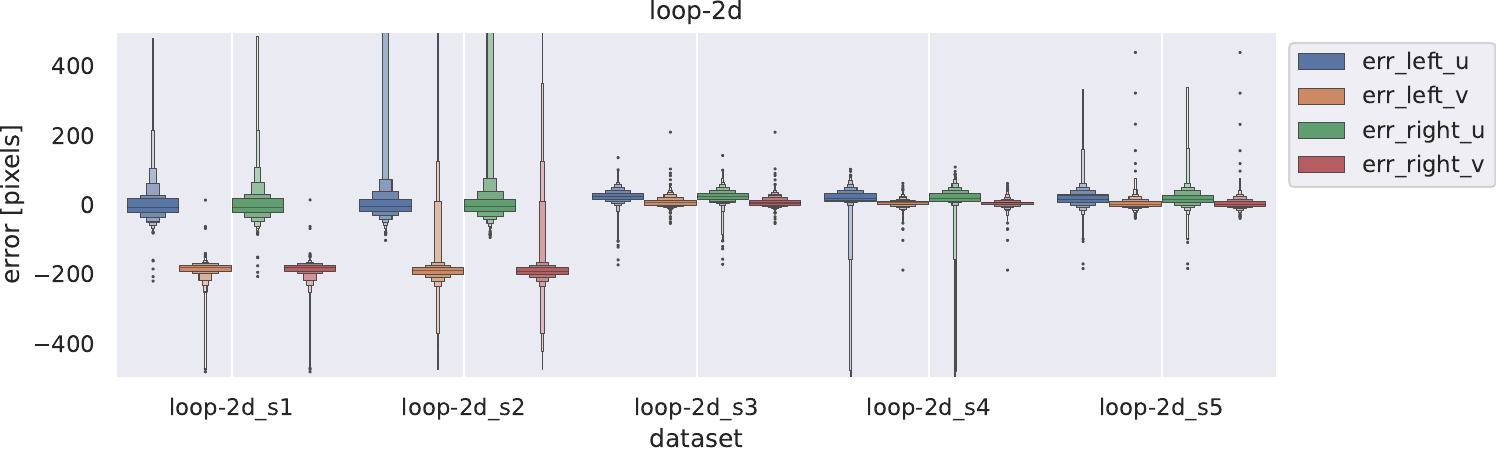}
  \includegraphics[height=3.9cm]{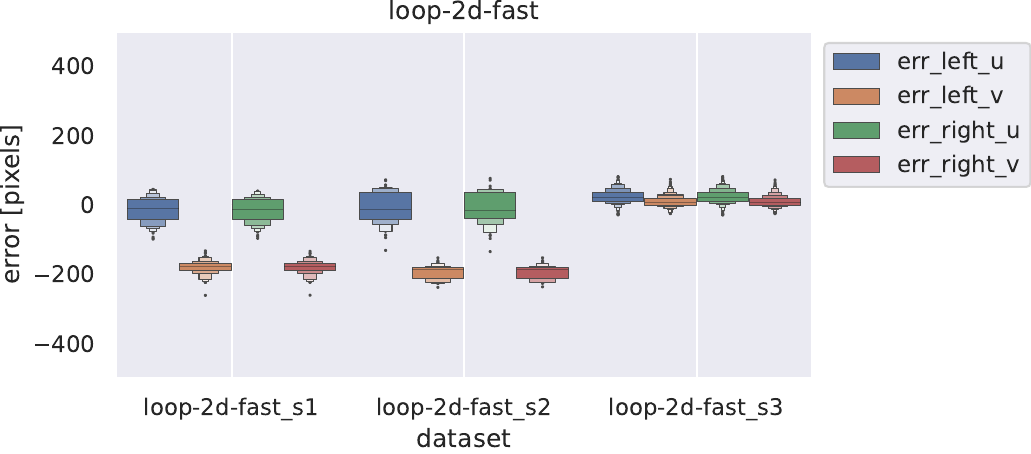}
  \includegraphics[height=3.9cm]{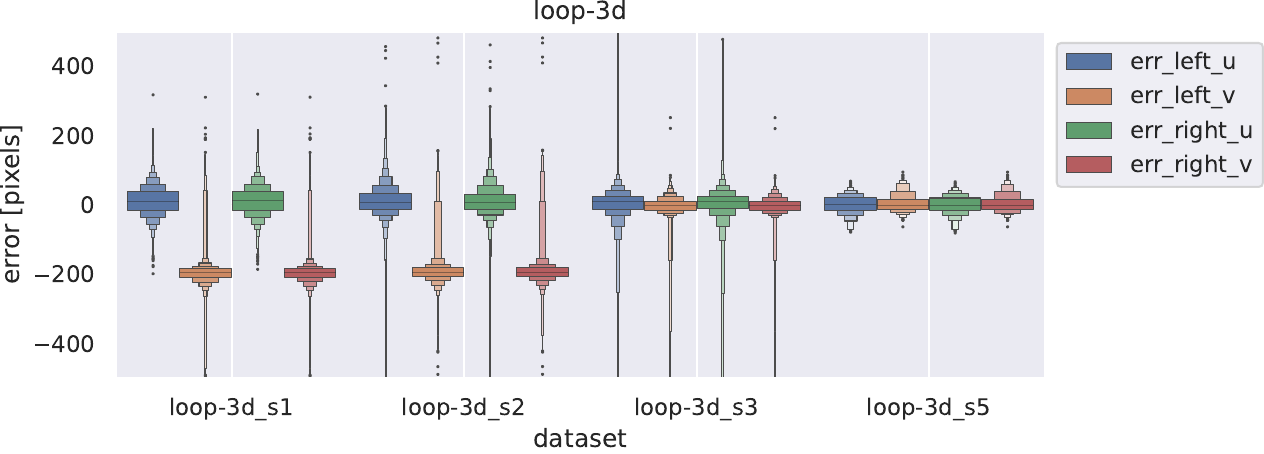}
  \includegraphics[height=3.9cm]{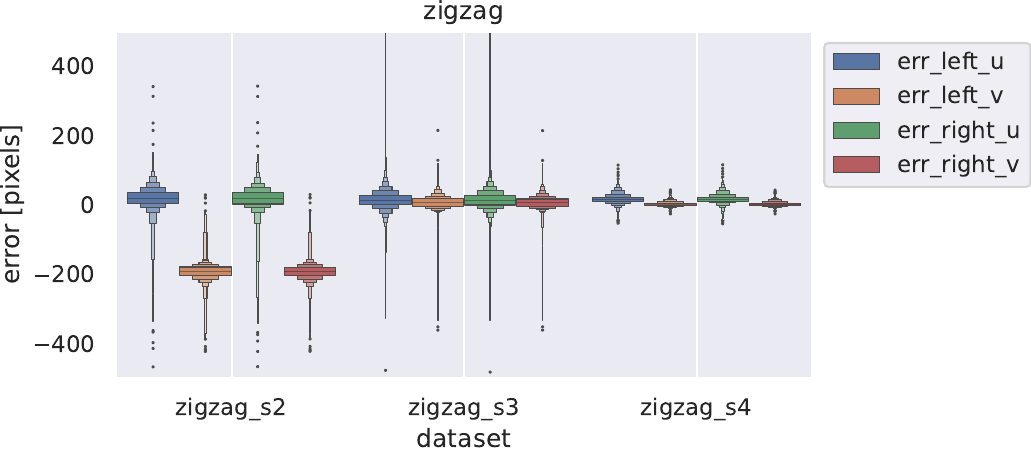}
  \includegraphics[height=3.9cm]{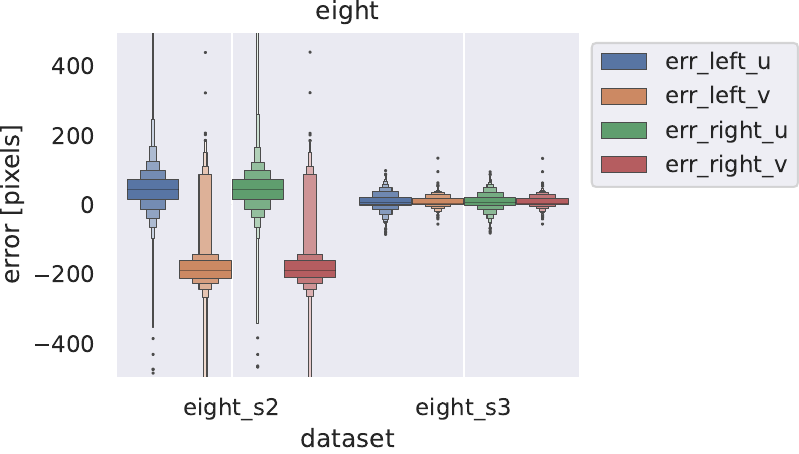}
  \includegraphics[height=3.9cm]{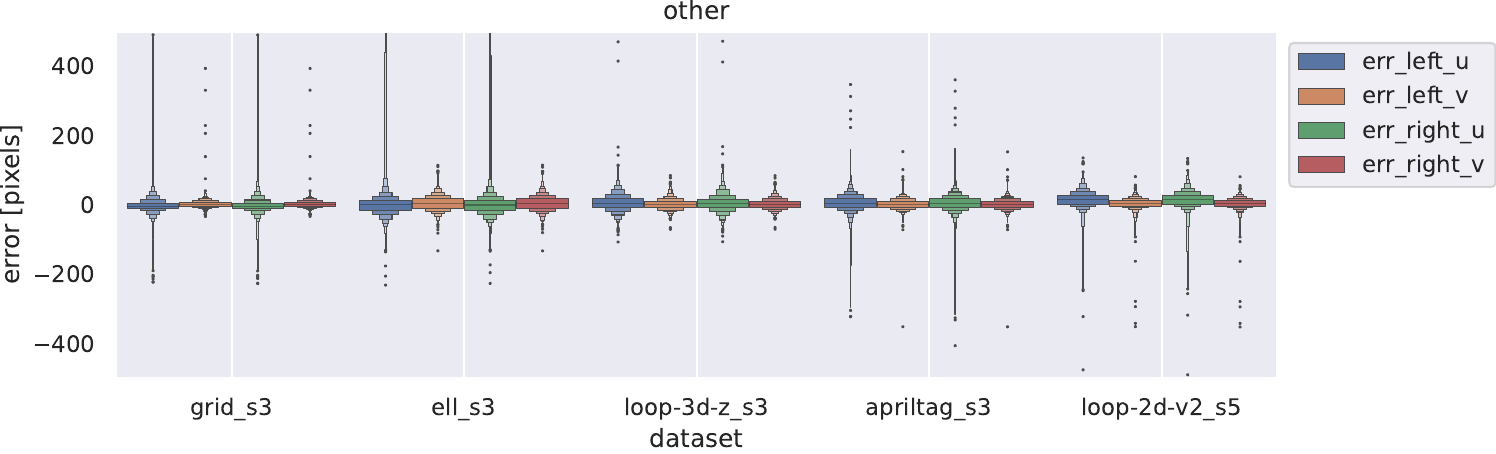}
  \caption{Stereo camera performance over all datasets.}
  \label{fig:overview-apriltag}
\end{figure}

The quality of stereo measurements per dataset is measured by the difference between the projected pixel values using the ground truth landmark locations and camera pose, and the measured pixel values, as shown in Figure~\ref{fig:overview-apriltag}. The measurements shown are \textit{before calibration}. 

\clearpage\FloatBarrier
\section{Data processing}

\subsection{Stereo camera calibration}

Stereo camera calibration involves two main steps:
\begin{enumerate}
  \item \textbf{Data association}: Involves associating \i{Apriltag} identifiers to the ground truth Vicon locations.
  \item \textbf{Extrinsic calibration}: Involves solving for the transformation between the Vicon frame attached to the camera rig and the left \i{Zed2i} camera frame (at focal point).
\end{enumerate}
Note that the rectified intrinsic parameters provided for the \i{Zed2i} camera were assumed to be accurate and were not tuned in this calibration procedure. 

Data association was performed using a camera sequence with \apriltag labels to manually label the set of Vicon ground truth landmark locations. 

For extrinsic calibration, we provide three different methodologies: fixed calibration (appendix \texttt{\_cal}), global calibration (\texttt{\_cal\_global})  and individual calibration (\texttt{\_cal\_individual}) For fixed calibration, the CAD model of the sensor rig was used to establish a transformation between Vicon and camera frames. For global calibration, we use the dataset \texttt{apriltag\_s3} to find the transformation that minimizes the reprojection (also called photometric) error using Gauss-Newton optimization~\cite{state-estimation}. For individual calibration, we use the same procedure, but calculate the best transformation for each dataset individually.

To reduce the effect of outliers, we perform the calibration three times, removing outliers after each iteration. Outliers are pixel measurements with errors higher than a fixed threshold (we used 100 for \i{v2} and \i{v3} and 300 for \i{v1}). 

\subsection{\ac{UWB} calibration}

\begin{figure}[h]
    \centering
    \includegraphics{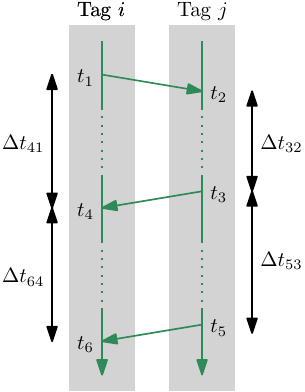}
    \caption{The DS-TWR protocol used in the experiments to generate range measurements between any two UWB tags.}
    \label{fig:ds_twr}
\end{figure}

The ranging protocol used in these experiments is the double-sided two way ranging (DS-TWR) protocol shown in Figure \ref{fig:ds_twr}. Each of the 6 timestamps $t_i, i \in \{1, \ldots, 6\}$ is recorded as part of the dataset, as well as the received signal power at the timestamps $t_2$ and $t_4$.

\begin{figure}[h]
  \centering
  \begin{minipage}{.48\linewidth} \centering
  \includegraphics[width=\linewidth]{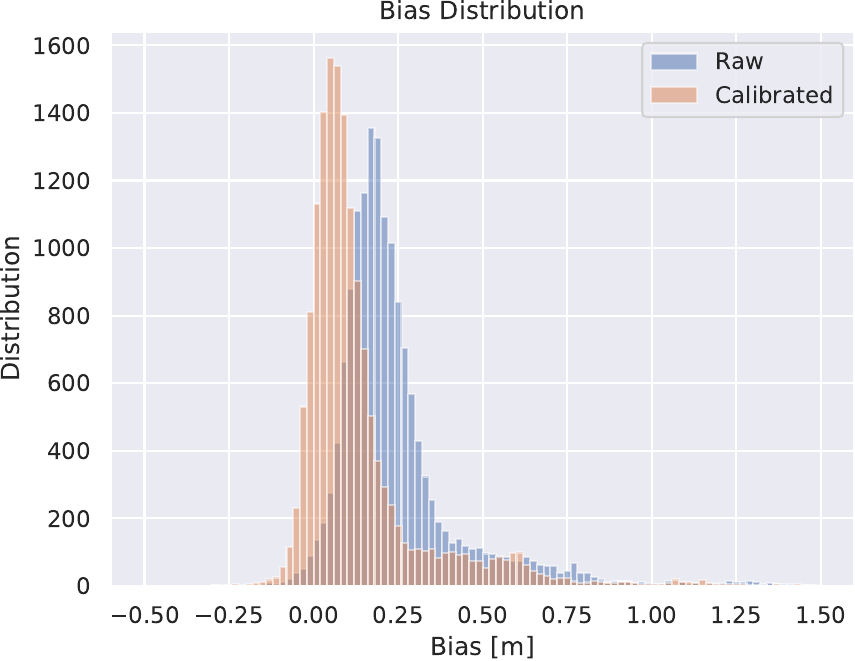}
  \caption{Histogram representing the distribution of the bias of the range measurements for the zigzag s2 experiment. The distribution for the raw and the calibrated measurements are shown.}
  \label{fig:uwb_histogram}
  \end{minipage} \hspace{5pt}
  \begin{minipage}{.48\linewidth} \centering
  \includegraphics[width=\linewidth]{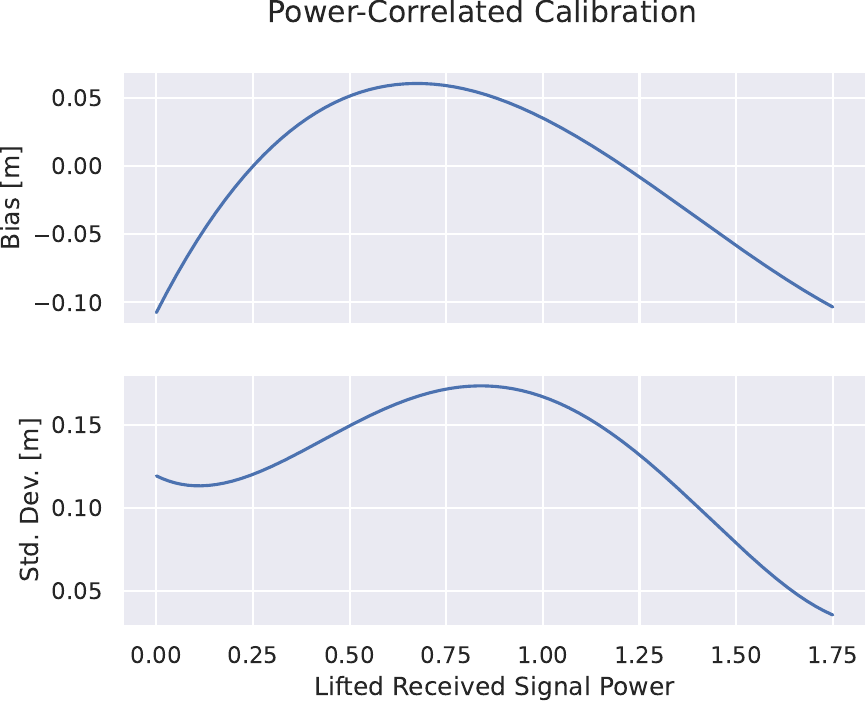}
  \caption{The relation between the bias and the standard deviation of the bias for the measurements as a function of the lifted received signal power. These are obtained from a separate calibration experiment.}
  \label{fig:power_calib}
  \end{minipage} 
\end{figure}

A time-of-flight measurement can be generated from the recorded timestamps as 
$$ t_f = \frac{1}{2} \left( \Delta t_{41} - \frac{\Delta t_{64}}{\Delta t_{53}} \Delta t_{32} \right),$$ where, as shown in Figure \ref{fig:ds_twr}, $\Delta t_{ji} = t_j - t_i$. Nonetheless, the range measurements are biased as shown in the blue distribution in Figure \ref{fig:uwb_histogram}. This error stems from multiple factors such as timestamping inaccuracies, multipath propagation, and obstacles. 

\begin{table}[h]
   \centering
   \caption{Calibrated antenna delays for all tags using a separate calibration experiment.}
   \label{tab:antenna_delay}
   \begin{tabular}{ c|c } 
    Tag ID & Antenna Delay [ns] \\ 
    \hline\hline
    1 & -0.5672 \\ 
    2 & -0.2139 \\ 
    3 & -0.2499 \\ 
    4 & -0.5514 \\ 
    5 & -0.3756 \\ 
    6 & -0.3292 \\ 
    7 & -0.1349 \\ 
    9 & -0.2914 \\ 
    10 & -0.3000 \\ 
    11 & -0.3953 \\
    12 & -0.3069 
   \end{tabular}
\end{table}

To obtain more accurate range measurements, the calibration procedure presented in~\cite{shalaby2023} is implemented. The antenna delay values shown in Table \ref{tab:antenna_delay} and the power-to-bias relation shown in Figure~\ref{fig:power_calib} are obtained by performing two separate experiments with 6 UWB tags fitted on 3 flying quadcopters (2 per quadcopter). A relation between the uncertainty or standard deviation of measurements as a function of the received signal power is also learned and is shown in the bottom plot of Figure \ref{fig:power_calib}. More information on the experiment and how these relations are obtained can be found in \cite{shalaby2023}.

The timestamps are then corrected using the antenna delays and the received-signal power and the range measurements are recomputed, giving much less biased measurements as shown in Figure \ref{fig:uwb_histogram}. Both the raw and calibrated measurements are provided when available, and they can be found under the columns \texttt{range} and \texttt{range\_calib}, respectively. The uncertainty of each measurement is also provided under the \texttt{std} column.

\section{Acknowledgments}

We would like to thank Joey Zou for his help in preparing the \textit{Decawave} data collection pipeline and rig adjustments. We also thank Connor Jong for his help during experiments and Abhishek Goudar for advice on \ac{UWB} data collection.

\newpage\FloatBarrier
\begin{appendices}
\section{Individual dataset plots}\label{app:data}

Each dataset folder in \texttt{data/<dataset\_name>} contains a number of plots for fast data inspection. The different plots are described in more detail below.

\foreach \filename in {%
loop-2d_s3%
}{%
\begin{figure}[h]
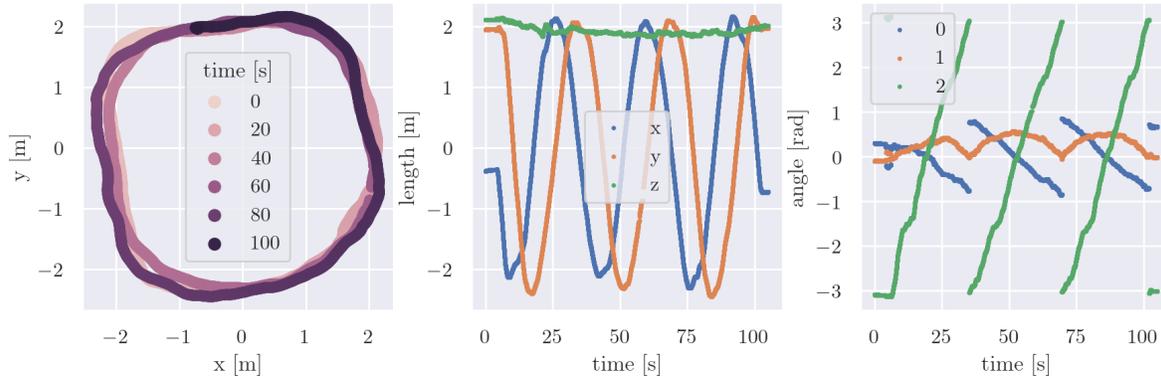

  \centering
    \includegraphics[width=.32\linewidth]{_figures/\filename/gt-xy.png}
    \includegraphics[width=.32\linewidth]{_figures/\filename/gt-t-xyz.png} 
    \includegraphics[width=.32\linewidth]{_figures/\filename/gt-t-angles.png}
    \caption{Ground truth data, from left to right: projection of trajectory to x-y plan (\texttt{gt-xy.png}), translation over time (\texttt{gt-t-xyz.png}) and rotation over time, using axis-angle representation (\texttt{gt-t-angles}).}
\end{figure}
\begin{figure}[h]
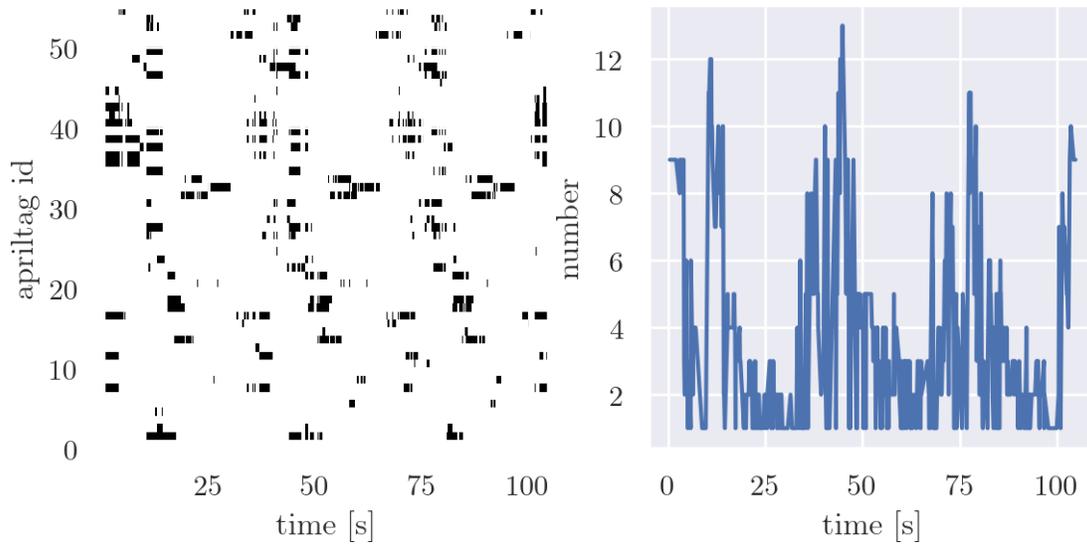

  \centering
   \includegraphics[width=.45\linewidth]{_figures/\filename/apriltag_mask.png}
   \includegraphics[width=.45\linewidth]{_figures/\filename/apriltag_number.png}
   \caption{Apriltag data, from left to right:  mask of \i{Apriltag} detections over time (\texttt{apriltag\_mask.png}), black corresponding to at least one detection in either the left or the right stereo camera frame, and number of synchronized \i{Apriltag} detections over time (\texttt{apriltag\_number.png}).}
\end{figure}

\begin{figure}[h]
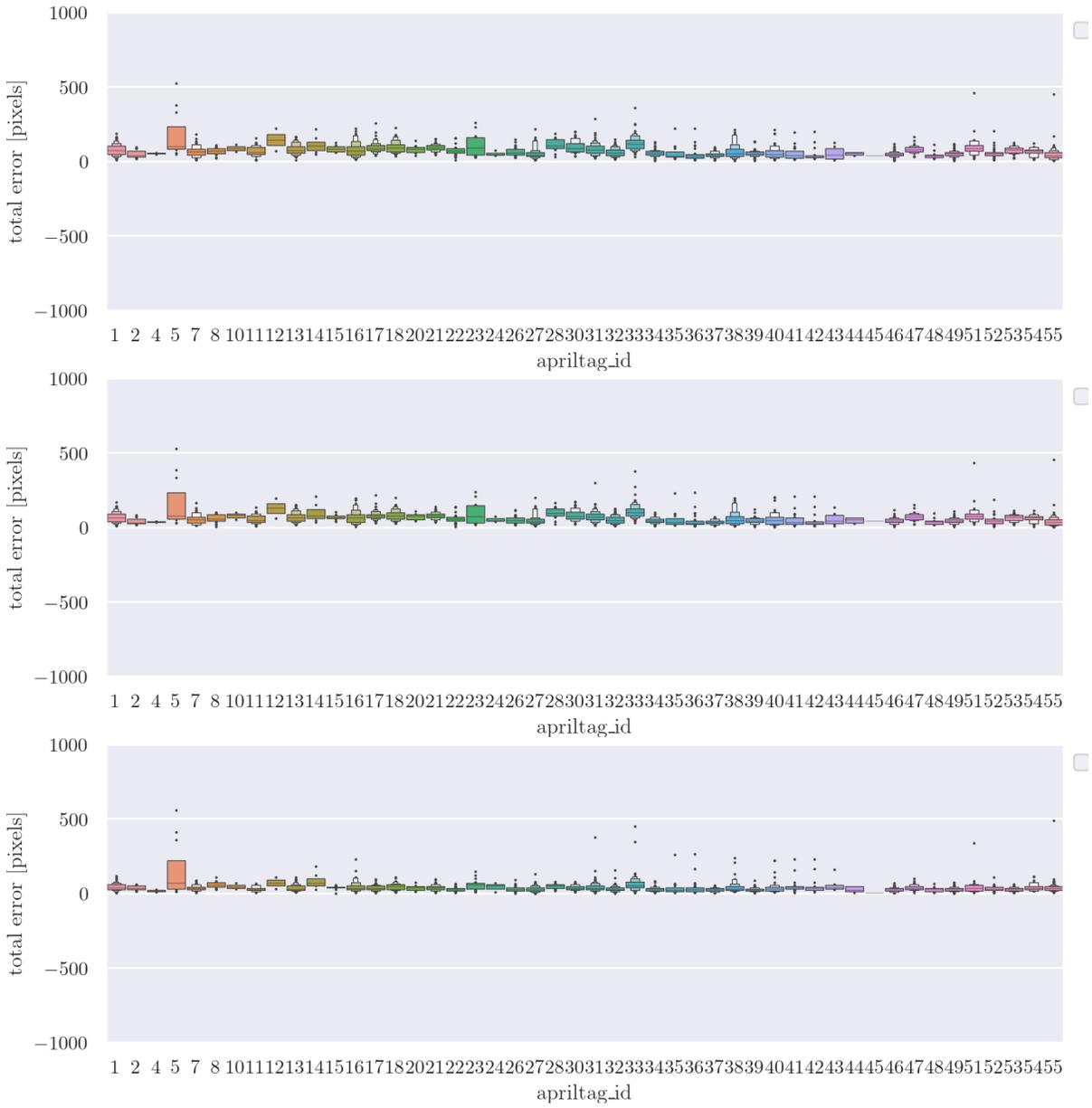

  \centering
  \includegraphics[width=\linewidth]{_figures/\filename/apriltag_cal.png}
  \includegraphics[width=\linewidth]{_figures/\filename/apriltag_cal_global.png}
  \includegraphics[width=\linewidth]{_figures/\filename/apriltag_cal_individual.png}
  \caption{Stereo camera pixel errors, using no calibration (top, \texttt{apriltag\_cal.pdf}), one global calibration (middle, \texttt{apriltag\_cal\_global.pdf}) and individual calibrations (bottom, \texttt{apriltag\_cal\_individual.pdf}). We also provide plots cropped at $\pm$ 300 pixels (\texttt{\_crop}).}
\end{figure}

\begin{figure}[h]
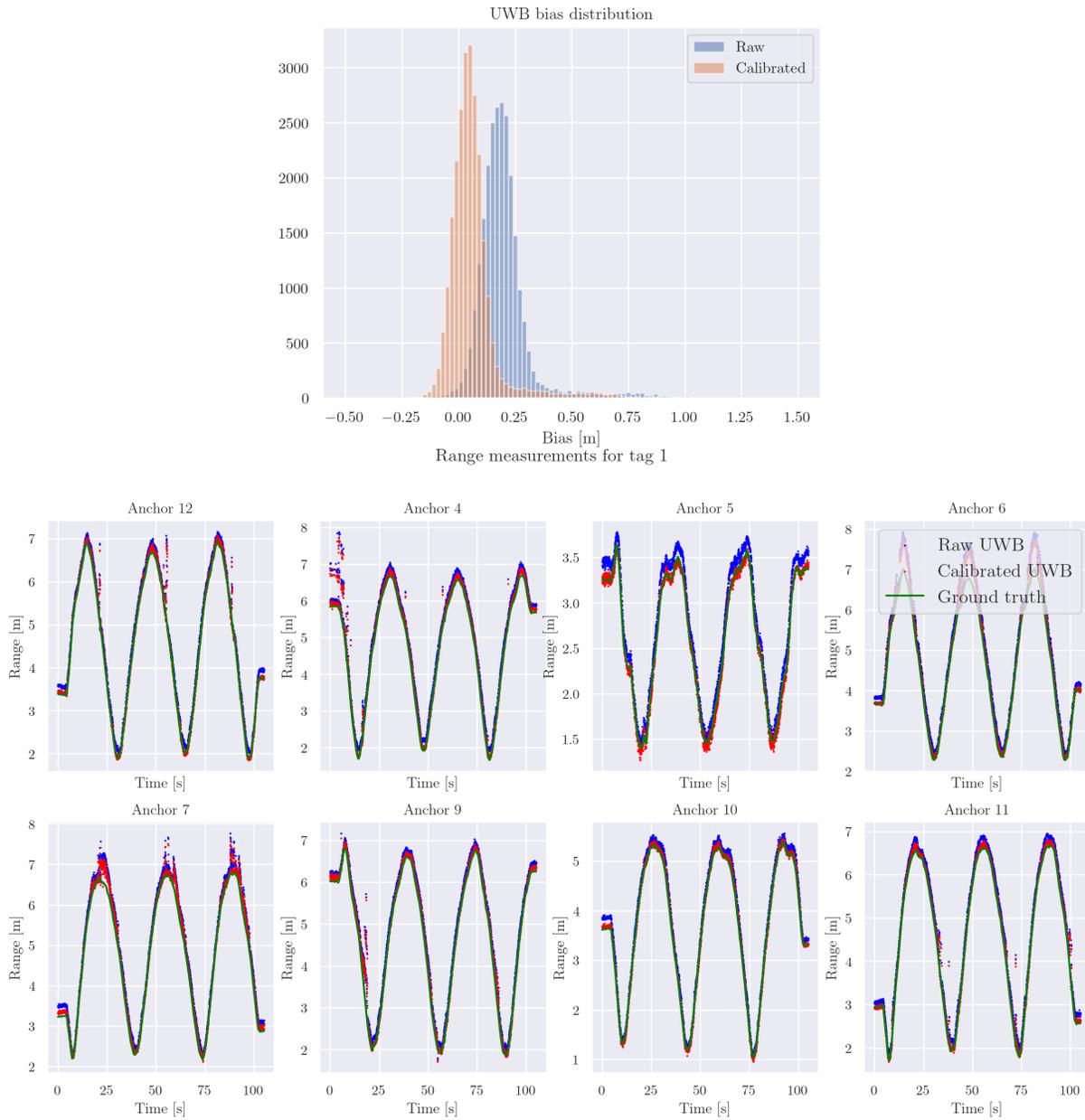

  \centering
  \includegraphics[width=.5\linewidth]{_figures/\filename/uwb_bias_histogram.png} \\
  \includegraphics[width=\linewidth]{_figures/\filename/uwb_measurements_tag_1.png}
  \caption{\ac{UWB} data, top: \ac{UWB} bias before and after calibration (\texttt{uwb\_bias\_histogram.png}), bottom: raw and calibrated measurements compared to ground truth (\texttt{uwb\_measurements\_tag\_1.png}).}
\end{figure}
}

\end{appendices}

\end{document}